\documentclass{article} 
\usepackage{iclr2025_conference,times}

\usepackage{amsmath,amsfonts,bm}

\def\eqref#1{equation~\ref{#1}}

\def\1{\bm{1}}

\DeclareMathAlphabet{\mathsfit}{\encodingdefault}{\sfdefault}{m}{sl}
\SetMathAlphabet{\mathsfit}{bold}{\encodingdefault}{\sfdefault}{bx}{n}

\usepackage{hyperref}
\usepackage{url}
\usepackage{graphicx}

\title{Preference-Based Alignment of Discrete\\Diffusion Models}

\author{Umberto Borso\textsuperscript{\textnormal{1,2}}\thanks{Work done at Centre for Artificial Intelligence, University College London. Correspondence to \texttt{uborso@student.ethz.ch}}, Davide Paglieri \textsuperscript{\textnormal{2}}, Jude Wells\textsuperscript{\textnormal{2}}, Tim Rocktäschel\textsuperscript{\textnormal{2}}  \\
\textsuperscript{1}ETH Zurich, \textsuperscript{2}Centre for Artificial Intelligence, University College London
}

\newcommand{\DthreePO}{\textbf{\texttt{D2-DPO}}}

\iclrfinalcopy 
\begin{document}

\maketitle

\begin{abstract}
Diffusion models \citep{ho2020denoising,song2020denoising} have achieved state-of-the-art performance across multiple domains \citep{austin2021structured,watson2023novo,anand2022protein}, with recent advancements extending their applicability to discrete data \citep{lou2023discrete,shi2024simplified,campbell2022continuous,campbell2024flow}. However, aligning discrete diffusion models with task-specific preferences remains challenging, particularly in scenarios where explicit reward functions are unavailable. In this work, we introduce \textbf{D}iscrete \textbf{D}iffusion \textbf{DPO} (\DthreePO{}), the first adaptation of Direct Preference Optimization (DPO) \citep{rafailov2024direct} to discrete diffusion models formulated as continuous-time Markov chains. Our approach derives a novel loss function that directly fine-tunes the generative process using preference data while preserving fidelity to a reference distribution. We validate \DthreePO{} on a structured binary sequence generation task, demonstrating that the method effectively aligns model outputs with preferences while maintaining structural validity. Our results highlight that \DthreePO{} enables controlled fine-tuning without requiring explicit reward models, making it a practical alternative to reinforcement learning-based approaches. Future research will explore extending \DthreePO{} to more complex generative tasks, including language modeling and protein sequence generation, as well as investigating alternative noise schedules, such as uniform noising, to enhance flexibility across different applications.

\end{abstract}

\section{Introduction}
Diffusion models have emerged as powerful generative models, achieving state-of-the-art results in a variety of domains, including image generation \citep{ho2020denoising,song2020denoising} and molecular design \citep{watson2023novo,anand2022protein}. While originally formulated in continuous spaces, recent advancements have extended diffusion models to discrete domains \citep{austin2021structured,campbell2022continuous}, including language modelling \citep{lou2023discrete,shi2024simplified,sahoo2024simple,ou2024your}, symbolic music composition \citep{campbell2022continuous} and biological sequence generation \citep{campbell2024flow}. Discrete diffusion models have demonstrated remarkable effectiveness in tasks where autoregressive approaches struggle, particularly in capturing long-range dependencies and modelling global consistency. However, in many applications, generating plausible sequences alone is insufficient. One often seeks to optimize generation with respect to specific task objectives, such as increasing factual accuracy in text generation, generating more harmonious music compositions, or designing protein sequences with improved stability.

To address this challenge, recent works have explored fine-tuning pre-trained discrete diffusion models to optimize task-specific reward functions \citep{wang2024fine}. However, explicitly defining a reward function is often infeasible when generation quality depends on subjective or hard-to-quantify criteria. In such cases, experts' feedback can provide valuable guidance: they can qualitatively compare generated candidates and express preferences based on fundamental knowledge of the domain.

Direct Preference Optimization (DPO) has recently emerged as a powerful method for fine-tuning generative models based on preference data, eliminating the need for explicit reward modelling. It has been successfully applied in natural language processing to align model responses with human feedback \citep{rafailov2024direct}, in text-to-image generation to improve adherence to human aesthetic preferences \citep{wallace2024diffusion}, and in protein design to enhance the stability of generated sequences \citep{widatalla2024aligning}. Despite its success in autoregressive and continuous generative models, DPO has not been explored for discrete diffusion models, which differ fundamentally in their formulation and training dynamics.


In this work, we introduce \textbf{D}iscrete \textbf{D}iffusion \textbf{DPO} (\DthreePO{}), the first adaptation of DPO to discrete diffusion models. Unlike continuous diffusion models, which leverage score-matching, discrete diffusion models are formulated as Continuous-Time Markov Chains (CTMCs), requiring a different optimization framework. We derive a novel loss function that directly fine-tunes discrete diffusion models using pairwise preference data while preserving fidelity to a reference distribution.

Our key contributions are as follows. Firstly, we introduce \DthreePO{}, a DPO-based optimization framework tailored for CTMCs, enabling preference alignment in discrete diffusion models without requiring an explicit reward function. Secondly, we show that under a masking-state noising process, our preference-based objective simplifies to an intuitive closed-form expression, providing theoretical insights into its effectiveness. Thirdly, we empirically validate \DthreePO{} on a structured sequence generation task, demonstrating that it successfully aligns discrete diffusion models with preferences while maintaining distributional coherence.

\section{Background and Notation}

\subsection{Discrete Diffusion Models}


\textbf{Continuous-Time Markov Chain (CTMC).} A CTMC describes a sequence of discrete states $\{x_t\}$ evolving over continuous time $t \in [0,1]$. The process begins at $t=0$ with an initial state $x_0 \sim p_0$, and transitions between states occur stochastically governed by a rate matrix $R_t \in \mathbb{R}^{\mathcal{X} \times \mathcal{X}}$. The probability of transitioning from state $x_t$ to $x_{t+dt}$ over an infinitesimal time interval $dt$ is given by:
\begin{align}
p_{t+dt | t}\left(x_{t+dt} | x_t\right) = \delta(x_t, x_{t+dt}) + R_t(x_t, x_{t+dt}) dt,
\label{eq:1}
\end{align}

where $\delta$ is the Kronecker delta function, which equals $1$ when $x_{t+dt} = x_t$ and $0$ otherwise. The off-diagonal elements of the rate matrix, $R_t(j, k) \geq 0$ for $j \neq k$, specify the rate at which probability mass transitions from state $j$ to state $k$ at time $t$. The diagonal elements $R_t(j, j) = -\sum_{k \neq j} R_t(j, k)$ represent the total rate at which probability mass moves out of state $j$ and are thus negative. 

\textbf{Noising Process.} 
The noising process $q_{t | 1}(x_t | x_1)$ progressively perturbs the data distribution $p_1(x) = p_{\text{data}}(x)$ gradually transforming it into the noise prior $p_0(x) = p_{\text{noise}}(x)$ as $t\to0$. A widely used approach is the masking-state noise process \citep{shi2024simplified, sahoo2024simple, ou2024your} which gradually maps all states $x\in\mathcal{X}$ to a masked state $M$ as $t \to 0$. Under this scheme, the noise prior is $p_{\text{noise}}^{\text{mask}}(x) = \delta\{M, x\}$, and the state space is augmented to $\mathcal{X} \cup \{M\}$. The corresponding transition kernel for this process is given by:
\begin{align}
q_{t | 1}^{\text{mask}}\left(x_t | x_1\right) = t \delta(x_1, x_t) + (1-t) \delta(M, x_t).
\label{eq: masking process}
\end{align}




\textbf{Generative Modelling.}
To generate samples from $p_{\text{data}}(x)$, we begin by drawing the initial noisy state from the noise prior, $x_0 \sim p_{\text{noise}}(x)$, and then simulate the trajectory $\{x_t\}_{t=0}^{t=1}$ by iteratively applying the transition kernel $p_{t+dt | t}(x_{t+dt} | x_t)$. This process allows the system to evolve towards the target distribution, ensuring that the final state at $t=1$ is effectively a sample from the clean data distribution, i.e., $x_1 \sim p_{\text{data}}(x)$.

Reconstructing the transition kernel in \eqref{eq:1} requires knowledge of the rate matrix $R_t(x_t, x_{t+dt})$. \citet{campbell2024flow} demonstrate that this matrix can be expressed as an expectation over a simpler conditional rate matrix. Specifically, we can write:
\begin{align}
R_t(x_t, x_{t+dt}) = \mathbb{E}_{p_{1 | t}(x_1 | x_t)} \left[ R^q_t(x_t, x_{t+dt} | x_1) \right],
\label{eq: rate matrix cond exp}
\end{align}

where $p_{1 | t}(x_1 | x_t)$ represents the  denoising distribution, which we approximate using a neural network $p_{1 | t}^\theta(x_1 | x_t)$. We define the rate matrix $R_t^\theta(x_t, x_{t+dt})$ by substituting  $p_{1 | t}^{\theta}(x_1 | x_t)$  into the expectation. The conditional rate matrix $R^q_t(x_t, x_{t+dt} | x_1)$ depends on the chosen noise schedule and is defined as:
\begin{align}
R^q_t(x_t, x_{t+dt} | x_1) = \frac{\operatorname{ReLU} \left( \partial_t q_{t | 1} (x_{t+dt} | x_1) - \partial_t q_{t | 1} (x_t | x_1) \right)}{S \cdot q_{t | 1} (x_t | x_1)}.
\label{eq: cond rate matrix relu}
\end{align}








\subsection{Direct Preference Optimization}

\textbf{Bradley-Terry (BT) Model.}
We assume access to a dataset of pairwise preferences $\mathcal{P}$ over clean data samples $x_1$. Each preference is represented as a tuple $(x_1^w, x_1^l, c)$, where $c \in \mathcal{C}$ represents a conditioning variable, $x_1^w$ is the preferred sample, and $x_1^l$ is the less preferred sample. The ranking between samples is assumed to follow an unknown latent reward function $r(c, x_1)$, such that $x_1^w \succ x_1^l \: \Longleftrightarrow \: r(c, x_1^w) > r(c, x_1^l)$. To model the probability of preferring $x_1^w$ over $x_1^l$, we adopt the Bradley-Terry (BT) model:
\begin{align}
p_{\operatorname{BT}}(x_1^w \succ x_1^l |c) = \sigma(r(c,x_1^w)-r(c,x_1^l)),
\end{align}

where $\sigma(\cdot)$ is the sigmoid function. Given a dataset of preferences, a parametric reward function can be learned by maximum likelihood estimation:
\begin{align}
L_{\mathrm{BT}}(\phi)=-\mathbb{E}_{c, x_1^w, x_1^l}\left[\log \sigma\left(r_\phi\left(c, x_1^w\right)-r_\phi\left(c, x_1^l\right)\right)\right].
\label{eq: BT loss}
\end{align}
\textbf{RLHF.}
Given a learned reward function $r_\phi(c, x_1)$, RLHF seeks to optimize a conditional generative model $p_\theta(x_1 | c)$ such that the expected reward is maximized while maintaining distributional regularization. The objective function takes the form:
\begin{align}
\max_{p_\theta} \mathbb{E}_{c \sim \mathcal{C}, x_1 \sim p_\theta\left(x_1 | c\right)}
\left[ r\left(c, x_1\right) \right]  -\beta \mathbb{D}_{\mathrm{KL}}\left[p_\theta\left(x_1 | c\right) \mid\mid p_{\mathrm{ref}}\left(x_1 | c\right)\right].
\label{eq: RLHF}
\end{align}
Here, $p_{\mathrm{ref}}(x_1 | c)$ is a reference model, and $\beta$ controls regularization.

\textbf{DPO.}
The optimizer of the RLHF objective in \eqref{eq: RLHF} can be written as:
\begin{align}
p_\theta(x_1|c) = p_{\operatorname{ref}}(x_1|c)\exp(r(c,x_1)/\beta)/Z(c),
\label{eq: rlhf optimizer}
\end{align}

where $Z(c) = \sum_{x_1} p_{\text{ref}}\left(x_1|c\right) \exp \left(r\left(c,x_1\right) / \beta\right)$ is a normalizing factor. Solving for $r(c,x_1)$ and substituting this into Equation \eqref{eq: BT loss}, we obtain the DPO loss function:
\begin{align}
L_{\mathrm{DPO}}(\theta)=-\mathbb{E}_{c, x_1^w, x_1^l}\left[\log \sigma\left(\beta \log \frac{p_\theta\left(x_1^w | c\right)}{p_{\text{ref}}\left(x_1^w | c\right)}-\beta \log \frac{p_\theta\left(x_1^l | c\right)}{p_{\text{ref}}\left(x_1^l | c\right)}\right)\right].
\label{eq:DPOloss}
\end{align}

This formulation eliminates the need for explicit reward modeling, allowing direct optimization of the generative model parameters $\theta$ without requiring an RL-based policy update.

\section{DPO for Discrete Diffusion Models}

To facilitate computations, we approximate the CTMC with a discrete-time representation. We partition the continuous time interval $[0,1]$ into equally spaced steps $t_n$ with $n \in \{0,...,N\}$, such that the process is described by a discrete-time Markov chain. Denoting the discrete-time states as $x_n=x_{t_n}$ we express the transition probabilities as
\begin{align}
p_{\theta} (x_{n+1}|x_n) = \delta(x_{n+1},x_n) + R^\theta_n(x_n, x_{n+1}) \Delta t.
\end{align}

Here, $R^\theta_n(x_n, x_{n+1})$ denotes the time-discretized rate matrix that governs state transitions. Building on the approach of \citet{wallace2024diffusion} we can express the DPO objective in discrete time $L_{\text{DT}}(\theta) =$
\begin{align}
-\log \sigma \left( \beta N \mathbb{E}_{\substack{n\sim \mathcal{U}\{0,N\}\\x_n^{w,l}\sim q(x_n|x_N^{w,l})\\ x_{n+1}^{w,l}\sim q(x_{n+1}|x_{n}^{w,l}, x_N^{w,l}) }} \left[ \log \frac{p_\theta(x_{n+1}^{w} | x_n^{w})}{p_{\text{ref}}(x_{n+1}^{w} | x_n^{w})} - \log \frac{p_\theta(x_{n+1}^l | x_n^l)}{p_{\text{ref}}(x_{n+1}^l | x_n^l)} \right] \right)
\end{align}
where $q(x_n | x_N)$ is the discrete time equivalent of $q_{t|1}(x_t | x_1)$, and $q(x_{n+1}|x_{n}, x_N)$ is the discrete time equivalent of \eqref{eq: conditional q}. We omit $c$ for compactness. 
By substituting the transition probability expansion for rate matrices, and taking the continuous-time limit ($N \to \infty$, $\Delta t \to 0$), the final \DthreePO{} loss for CTMCs is obtained:
\begin{align}
    L_{\DthreePO{}}(\theta) = -\mathbb{E}_{\substack{(x_1^w, x_1^l) \sim \mathcal{P},t \sim \mathcal{U}[0,1] \\x^w\sim q(x_t|x_1^w), x^l\sim q(x_t|x_1^l) }}
    \log \sigma\Bigg[ \beta\:\mathcal{D}^{\theta}_\text{ref}(x_t^w|x_1^w) - \beta\:\mathcal{D}_\text{ref}^{\theta}(x_t^l|x_1^l) \Bigg]
    \label{eq: final ct loss}
\end{align}

with
\begin{align}
    \mathcal{D}^{\theta}_\text{ref}(x_t|x_1) = \sum_{j\neq x_t}R_t^{q}(x_t,j|x_1) \log \frac{ R_t^{\theta}(x_t,j)}{R_t^{\text{ref}}(x_t,j)} + R_t^{\text{ref}}(x_t,j)- R_t^{\theta}(x_t,j)\: ,
\end{align}
where $R_t^{q}(x_t,x_{t+dt}|x_1)$ depends on the chosen noise schedule and is defined as per \eqref{eq: cond rate matrix relu}, while $R_t^{\theta}(x_t,x_{t+dt})$ and $R_t^{\text{ref}}(x_t,x_{t+dt})$ are estimated as per \eqref{eq: rate matrix cond exp}. We defer the full derivation to Appendix \ref{sec: full derivation 1D} and the multi-dimensional case to Appendix \ref{sec: multidim dpo derivation}. In Appendix \ref{sec: dpo for masking state}, we show how this objective can be efficiently optimized for the masking-state noise process.
\section{Preliminary Experiments}\label{sec: experiments}
\begin{figure}[h]
\begin{center}
\includegraphics[width=\linewidth]{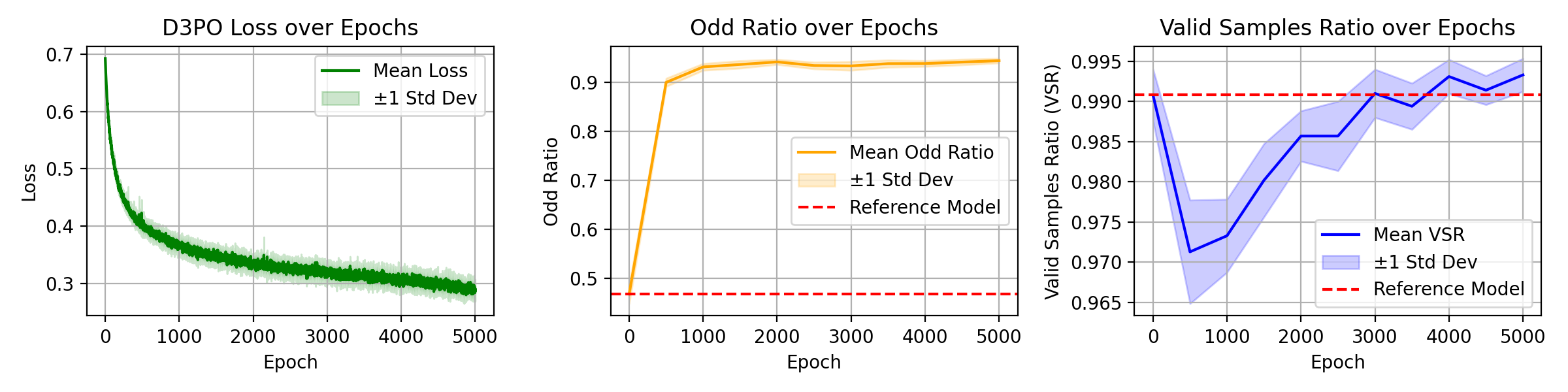}
\end{center}
\caption{Results for preference-based alignment using the \DthreePO{} loss. (Left) Training loss monotonically decreases over epochs. (Center) Ratio of generated sequences corresponding to odd integers increases w.r.t. reference model. (Right) Fraction of generated sequences with valid structure remains close to 1.}
\label{fig:d3po_aggregated_results}
\end{figure}

To validate the effectiveness of \DthreePO{}, we conduct a small-scale experiment demonstrating how the proposed loss in Equation \eqref{eq: final ct loss} enables preference alignment in a discrete diffusion model. Building on the framework of \citet{campbell2024flow}, we first pre-train a masking-state discrete diffusion model to generate structured binary representations of integers. Specifically, each integer $i \in \{0, \dots, N\}$  is represented as a binary sequence of length $N$, denoted as $b_i \in {0,1}^N$. The first $i$ bits are set to $1$, while the remaining bits are set to $0$. The pre-trained model learns to generate valid sequences that adhere to this structured encoding rather than producing arbitrary binary strings.

We then fine-tune the model using our preference-based objective in \eqref{eq: final ct loss} to bias the generative distribution toward binary sequences that represent odd integers. To achieve this, we construct a dataset of pairwise preferences, where the preferred sample $x^w$ corresponds to an odd integer and the less preferred sample $x^l$ corresponds to an even integer. Figure \ref{fig:d3po_aggregated_results} summarizes the fine-tuning process. On the left, the training loss steadily decreases, indicating stable optimization. In the centre, the odd-integer ratio,proportion of generated sequences corresponding to odd integers, rapidly rises above 0.9, confirming model successfully shifts its generative distribution toward odd numbers. On the right, the Valid Samples Ratio (VSR) measures the fraction of generated sequences that correctly follow the structured binary encoding of integers. After an initial dip, the VSR steadily recovers and surpasses the reference baseline, confirming that fine-tuning does not compromise structural validity.

\section{Conclusion and Future Work}
We introduce Discrete Diffusion DPO (\DthreePO{}), a novel extension of the DPO framework to diffusion models formulate as continuous-time Markov chains. Our derivation yields a computationally efficient loss function that aligns the generative sampling process with preference data while preserving fidelity to the reference distribution. Experiments on a structured binary sequence generation task confirmed that \DthreePO{} successfully biases discrete diffusion models towards preferred outputs while preserving structural validity.

Future work will explore scalability to larger models and more complex sequence generation tasks, such as language modelling and protein design. Additionally, we aim to investigate alternative noise schedules, including the uniform noise schedule, where the prior is a uniform distribution over states, potentially enhancing flexibility in different applications. 

\bibliography{sections/references}
\bibliographystyle{iclr2025_conference}

\appendix
\section{Appendix Structure}
The appendix is structured as follows. Appendix \ref{sec: related work} discusses related work, covering advancements in discrete diffusion models, fine-tuning techniques, and preference-based optimization in diffusion models. Appendix \ref{sec: full derivation 1D} provides a detailed derivation of the \DthreePO{} loss for discrete diffusion models, starting from a discrete-time approximation of the CTMC formulation and extending it to the continuous-time limit. Appendix \ref{sec: multidim dpo derivation} generalizes the \DthreePO{} loss to multi-dimensional data, presenting a factorized transition model that enables tractable optimization in structured sequence generation tasks. Appendix \ref{sec: dpo for masking state} derives the \DthreePO{} loss for the masking noise process, adapting the framework for discrete diffusion models that use an absorbing-state corruption scheme. Appendix \ref{sec: uniform noise masking} extends the masking noise derivation to cases with additional re-masking noise, allowing for bidirectional transitions between masked and unmasked states. Appendix \ref{sec: complexity masking} provides a complexity analysis of the derived loss functions for the masking state noise process, showing that preference-based fine-tuning with \DthreePO{} is computationally efficient.

\section{Related Work} \label{sec: related work}

\textbf{Discrete Diffusion Models.} Diffusion models have achieved strong generative performance in continuous spaces \citep{ho2020denoising,song2020denoising}, with recent extensions to discrete spaces enabling applications in language modelling and biological sequence design \citep{austin2021structured,campbell2022continuous,lou2023discrete,shi2024simplified,sahoo2024simple,ou2024your}. Compared to autoregressive models, discrete diffusion models better capture long-range dependencies and generate structured sequences such as DNA and protein sequences \citep{sarkar2024designing,campbell2024flow}.

\textbf{Fine-Tuning and Alignment of Discrete Diffusion Models.} Fine-tuning diffusion models for controlled generation typically involves guidance techniques, RL-based optimization, or classifier-free methods. Guidance methods such as classifier-based guidance \citep{dhariwal2021diffusion,song2020denoising} have been extended to discrete spaces \citep{nisonoff2024unlocking}, but require costly iterative inference. RL-based fine-tuning has been explored for optimizing reward functions in continuous diffusion models \citep{fan2024reinforcement,black2023training} and discrete diffusion models \citep{wang2024fine}. Classifier-free fine-tuning \citep{ho2022classifier,zhang2023adding} conditions on high-reward samples, but is limited by reward sparsity in structured sequence generation. Our work departs from these approaches by proposing preference-based fine-tuning for discrete diffusion models, enabling optimization without an explicit reward model.

\textbf{Preference-Based Alignment of Diffusion Models.} Preference-based optimization methods such as Reinforcement Learning from Human Feedback (RLHF) \cite{ziegler2019fine} and Direct Preference Optimization (DPO) \citep{rafailov2024direct} have been highly effective for fine-tuning LLMs and continuous diffusion models. Unlike RL-based methods, DPO directly fine-tunes a model using pairwise preference comparisons, bypassing the need for a reward model \citep{ethayarajh2024kto,azar2024general}. Recent adaptations of DPO to text-to-image diffusion models \citep{zhu2025dspo,wallace2024diffusion,yang2024using,li2024aligning} have shown promising results but are not applicable to discrete diffusion models.

Our work extends DPO to discrete diffusion models, deriving a loss function that respects their underlying CMTC formulation. This enables preference-based fine-tuning without the need of a reward model.
\section{Full Derivation of 1-Dimensional \DthreePO{} Loss} \label{sec: full derivation 1D}
\subsection{Conditional Denoising Kernel.} 
Here we provide an expression for the infinitesimal transition probability $q_{t+dt | t,1}(x_{t+dt} | x_t,x_1)$ in terms of the conditional rate matrix $R^q_t(x_t, x_{t+dt} | x_1)$ which will be useful later in the derivation of the \DthreePO{} loss.

Given a noise process $q_{t | 1}(x_t | x_1)$ we can define the joint probability over two successive states $x_t$ and $x_{t+dt}$ as $q_{t,t+dt | 1}(x_t,x_{t+dt} | x_1)$. Using the chain rule of probability:
\begin{align*}
    q_{t,t+dt | 1}(x_t,x_{t+dt} | x_1) = q_{t | 1}(x_t | x_1)q_{t+dt | t,1}(x_{t+dt} |x_t, x_1)
\end{align*}

where $q_{t+dt | t,1}(x_{t+dt} |x_t, x_1)$ can be interpreted as an infinitesimal denoising probability, conditioned on clean data $x_1$. Similarly to \eqref{eq:1}, we can write this infinitesimal transition probability in terms of a rate matrix:
\begin{align}
    q_{t+dt | t,1}(x_{t+dt} | x_t,x_1) = \delta(x_t, x_{t+dt}) + R^q_t(x_t, x_{t+dt} | x_1)dt
    \label{eq: conditional q}
\end{align}

where the conditional rate matrix $R^q_t(x_t, x_{t+dt} | x_1)$ is given as per \eqref{eq: cond rate matrix relu}.

\subsection{Discrete-Time Approximation}
We consider a time-discretization of the CTMC to simplify calculations. In practice, we approximate the time evolution of the sequence trajectory $\{x_t\}$ using discrete steps of size  $\Delta t$, and successively take the limit as $\Delta t \to 0$ to recover the continuous time case. We partition the the time interval $[0,1]$ with discrete time steps $t_n, \:n\in \{0,...,N\}$ where $t_0=0$ and $t_{N}=1$. We define $\Delta t = t_n - t_{n-1} = 1/(N+1)$ hence recovering the continuous time case when $N\to\infty$. With a slight abuse of notation we write $x_n=x_{t_n}$.

Considering a CTMC with this time partitioning converts the problem into a discrete time Markov Chain with transition kernel $p_{\theta} (x_{n+1}|x_{n})$ which is the time-discrete equivalent of $p_{t+dt|t}^{\theta} (x_{t+dt}\mid x_t)$ that naturally emerges from \eqref{eq:1} by identifying $dt=\Delta t$ and evaluating at $t=t_n$. Hence we have:
\begin{align}
    p_{\theta} (x_{n+1}|x_{n}) &:= p_{t_n+\Delta t|t_n}^{\theta} (x_{t_n+\Delta t}|x_{t_n})\\
    &= \delta(x_{n+1},x_{n}) + R^\theta_n(x_n, x_{n+1}) \Delta t
    \label{eq:discrete1}
\end{align}
Following the Markov assumption we can factorize the joint probability over paths in discrete time
\begin{align}
p_\theta(x_{0:N})= p_\theta(x_0)\prod_{n=1}^Np_\theta(x_{n+1}|x_n).
\end{align}
We define $\mathcal{R}_{\text{DT}}(c, x_{0:N} )$ as the reward on the whole trajectory in discrete time, such that we can define $r_{\text{DT}}(c, x_1)$ as:
\begin{align}
    r_{\text{DT}}(c, x_{N}) &= \mathbb{E}_{x_{0:N-1}\sim  p_\theta\left(x_{0: N-1} \mid x_N, c\right) }\mathcal{R}_{\text{DT}}\left(c, x_{0: N}\right)
\end{align}

\subsection{RLHF loss for discrete diffusion models}
Now our derivation proceeds along the lines of \citet{wallace2024diffusion}, who derive a DPO loss function for classical diffusion models in discrete time. The RLHF objective in Eq. \eqref{eq: RLHF} can be adapted to the diffusion framework as:
\begin{align*}
&\max _{p_\theta} \mathbb{E}_{x_N \sim p_\theta\left(x_N \mid c\right)
}[ \left.r_{\text{DT}}\left(c, x_N\right)\right]  -\beta \mathbb{D}_{\mathrm{KL}}\left[p_\theta\left(x_N \mid c\right) \| p_{\mathrm{ref}}\left(x_N \mid c\right)\right] \\
=& \min_{p_\theta} - \mathbb{E}_{x_N \sim p_\theta\left(x_N \mid c\right)}[ \left.r_{\text{DT}}\left(c, x_N\right)\right]  +\beta \mathbb{D}_{\mathrm{KL}}\left[p_\theta\left(x_N \mid c\right) \| p_{\mathrm{ref}}\left(x_N \mid c\right)\right] \\
\leq & \min_{p_\theta} - \mathbb{E}_{x_N \sim p_\theta\left(x_N \mid c\right)}[ \left.r_{\text{DT}}\left(c, x_N\right)\right]  +\beta \mathbb{D}_{\mathrm{KL}}\left[p_\theta\left(x_{0:N}\mid c\right) \| p_{\mathrm{ref}}\left(x_{0:N} \mid c\right)\right] \\
= & \min_{p_\theta} - \mathbb{E}_{x_{0:N} \sim p_\theta\left(x_{0:N} \mid c\right)}[ \left.\mathcal{R}_{\text{DT}}\left(c, x_{0:N}\right)\right]  +\beta \mathbb{D}_{\mathrm{KL}}\left[p_\theta\left(x_{0:N}\mid c\right) \| p_{\mathrm{ref}}\left(x_{0:N} \mid c\right)\right] \\
=& \min_{p_\theta} - \mathbb{E}_{x_{0:N} \sim p_\theta\left(x_{0:N} \mid c\right)}[ \left.\mathcal{R}_{\text{DT}}\left(c, x_{0:N}\right)\right]  +\beta  \mathbb{E}_{x_{0:N} \sim p_\theta\left(x_{0:N} \mid c\right)} \left[\log \frac{p_\theta\left(x_{0:N}\mid c\right)}{p_{\mathrm{ref}}\left(x_{0:N} \mid c\right)}\right]\\
=&\min_{p_\theta} \mathbb{E}_{x_{0:N} \sim p_\theta\left(x_{0:N} \mid c\right)} \left[ \log\frac{p_\theta\left(x_{0:N}\mid c\right)}{p_{\mathrm{ref}}\left(x_{0:N} \mid c\right) \exp(\mathcal{R}_{\text{DT}}\left(c, x_{0:N}\right)/\beta)} \right] \\
=&\min_{p_\theta} \mathbb{E}_{x_{0:N} \sim p_\theta\left(x_{0:N} \mid c\right)} \left[ \log\frac{p_\theta\left(x_{0:N}\mid c\right)}{p_{\mathrm{ref}}\left(x_{0:N} \mid c\right) \exp(\mathcal{R}_{\text{DT}}\left(c, x_{0:N}\right)/\beta)/Z(c)} + \log Z(c) \right] \\
=&\min_{p_\theta} \mathbb{D}_{\mathrm{KL}} \left[p_\theta\left(x_{0:N}\mid c\right) || p_{\mathrm{ref}}\left(x_{0:N} \mid c\right) \exp(\mathcal{R}_{\text{DT}}\left(c, x_{0:N}\right)/\beta)/Z(c) \right]
\label{eq: RLHF DD}
\end{align*}

where $c \sim \mathcal{C}$, and on the third line we used the joint KL-divergence $\mathbb{D}_{\mathrm{KL}}[p_\theta(x_{0:N}\mid c) \| p_{\mathrm{ref}}(x_{0:N} \mid c)]$ as upper bound of the marginal $\mathbb{D}_{\mathrm{KL}}[p_\theta(x_N \mid c) \| p_{\mathrm{ref}}(x_N \mid c)]$. The unique global solution to this optimisation problem is given by:
\begin{align*}
    p_\theta^*\left(x_{0:N}\mid c\right) = p_{\mathrm{ref}}\left(x_{0:N} \mid c\right) \exp(\mathcal{R}_{\text{DT}}\left(c, x_{0:N}\right)/\beta)/Z(c) \quad, 
\end{align*}

Hence we can re-parametrize the reward function as:
\begin{align*}
\mathcal{R}_{\text{DT}}\left(c, x_{0: N}\right)=\beta \log \frac{p_\theta^*\left(x_{0: N} \mid c\right)}{p_{\text{ref}}\left(x_{0: N} \mid c\right)}+\beta \log Z(c)
\end{align*}

which leads to:
\begin{align}
r_{\text{DT}}(c, x_{N}) &= \mathbb{E}_{x_{0:N-1}\sim  p_\theta\left(x_{0: N-1} \mid x_N, c\right) }\mathcal{R}_{\text{DT}}\left(c, x_{0: N}\right)\\
&=\beta\mathbb{E}_{x_{0:N-1}\sim p_\theta\left(x_{0: N-1} \mid x_N, c\right) }\left[ \log \frac{p_\theta^*\left(x_{0: N} \mid c\right)}{p_{\text{ref}}\left(x_{0: N} \mid c\right)} \right]+\beta \log Z(c)
\label{eq: discrete reward}
\end{align}

\subsection{\DthreePO{} Loss}
We can substitute \eqref{eq: discrete reward} into the BT model loss in \eqref{eq: BT loss} to get the \textit{per-example} DPO loss in the discrete time approximation: 
\begin{align*}
    L_{\mathrm{DT}}(\theta)&= -\log \sigma\left(\beta\mathbb{E}_{\substack{x_{0:N-1}^w\sim p_\theta(x_{0: N-1}^w \mid x_N^w)\\\;x_{0:N-1}^l\sim p_\theta(x_{0: N-1}^l \mid x_N^l) }} \left[ \log\frac{p_\theta(x_{0: N}^w )}{p_{\text{ref}}(x_{0: N}^w )} - \log \frac{p_\theta(x_{0: N}^l )}{p_{\text{ref}}(x_{0: N}^l )} \right]\right) \\
    &= -\log \sigma\left(\beta\mathbb{E}_{\substack{x_{0:N-1}^w\sim p_\theta(x_{0: N-1}^w \mid x_N^w)\\\;x_{0:N-1}^l\sim p_\theta(x_{0: N-1}^l \mid x_N^l) }} \left[ \sum_{n=0}^{N-1} \log\frac{p_\theta(x_{n+1}^w|x_n^w )}{p_{\text{ref}}(x_{n+1}^w|x_n^w )} - \log \frac{p_\theta(x_{n+1}^l|x_n^l  )}{p_{\text{ref}}(x_{n+1}^l|x_n^l  )} \right]\right) \\
    &= -\log \sigma\left(\beta\mathbb{E}_{\substack{x_{0:N-1}^w\sim p_\theta(x_{0: N-1}^w \mid x_N^w)\\\;x_{0:N-1}^l\sim p_\theta(x_{0: N-1}^l \mid x_N^l) }} N\mathbb{E}_n\left[\log\frac{p_\theta(x_{n+1}^w|x_n^w )}{p_{\text{ref}}(x_{n+1}^w|x_n^w )} - \log \frac{p_\theta(x_{n+1}^l|x_n^l  )}{p_{\text{ref}}(x_{n+1}^l|x_n^l  )} \right]\right) 
\end{align*}
where we omit $c$ for simplicity. Since sampling from the reverse process $p_\theta(x_{0: N-1} \mid x_N)$ is intractable, we approximate it with the forward process $q(x_{0: N-1} \mid x_N)$: 
\begin{align*}
    L_{\mathrm{DT}}(\theta) &= -\log \sigma\left(\beta\mathbb{E}_{\substack{x_{0:N-1}^w\sim q(x_{0: N-1}^w \mid x_N^w)\\\;x_{0:N-1}^l\sim q(x_{0: N-1}^l \mid x_N^l) }} N\mathbb{E}_n\left[\log\frac{p_\theta(x_{n+1}^w|x_n^w )}{p_{\text{ref}}(x_{n+1}^w|x_n^w )} - \log \frac{p_\theta(x_{n+1}^l|x_n^l  )}{p_{\text{ref}}(x_{n+1}^l|x_n^l  )} \right]\right) \\
    &= -\log \sigma\left(\beta N\mathbb{E}_n\mathbb{E}_{\substack{x_{n+1,n}^w\sim q(x_{n+1,n}|x_N^w) \\\;x_{n+1,n}^l\sim q(x_{n+1,n}|x_N^l)  }} \left[\log\frac{p_\theta(x_{n+1}^w|x_n^w )}{p_{\text{ref}}(x_{n+1}^w|x_n^w )} - \log \frac{p_\theta(x_{n+1}^l|x_n^l  )}{p_{\text{ref}}(x_{n+1}^l|x_n^l  )} \right]\right)
\end{align*}

Using the chan rule we write $ q(x_{n+1,n}|x_N) = q(x_n|x_N)q(x_{n+1}|x_n,x_N)$, where $q(x_n | x_N)$ is the discrete time equivalent of $q_{t|1}(x_t | x_1)$, and $q(x_{n+1}|x_{n}, x_N)$ is the discrete time equivalent of \eqref{eq: conditional q}. Hence we get:
\begin{align}
    L_{\mathrm{DT}}(\theta) = -\log \sigma\Bigg(\beta N\mathbb{E}_n & \mathbb{E}_{\substack{x_n^w\sim q(x_n|x_N^w)\\ x_{n+1}^w\sim q(x_{n+1}|x_{n}^w, x_N^w)}} \left[\log\frac{p_\theta(x_{n+1}^w|x_n^w )}{p_{\text{ref}}(x_{n+1}^w|x_n^w )} \right] \notag\\
    -& \mathbb{E}_{\substack{x_n^l\sim q(x_n|x_N^l) \\  x_{n+1}^l\sim q(x_{n+1}|x_{n}^l,x_N^l)}} \left[\log \frac{p_\theta(x_{n+1}^l|x_n^l  )}{p_{\text{ref}}(x_{n+1}^l|x_n^l  )} \right]\Bigg)
    \label{eq: mid step derivation 1d}
\end{align}

Following \citet{campbell2022continuous} we will expand the expression for $\mathbb{E}_{ x_{n+1}\sim q(x_{n+1}|x_{n},x_N)} 
  \left[\log\frac{p_\theta(x_{n+1}|x_n )}{p_{\text{ref}}(x_{n+1}|x_n )} \right]$ starting from $\log p_\theta(x_{n+1}|x_n )$:
\begin{align*}
    \log p_\theta(x_{n+1}|x_n ) &= \log (\delta_{x_n,x_{n+1}} + R_n^\theta(x_n,x_{n+1})\Delta t) \\
    &= \delta_{x_n,x_{n+1}}\log (1 + R^\theta_n(x_n,x_{n})\Delta t) + (1-\delta_{x_n,x_{n+1}}) \log (R^\theta_n(x_n,x_{n+1})\Delta t) \\
    &= \delta_{x_n,x_{n+1}}R^\theta_n(x_n,x_{n})\Delta t + (1-\delta_{x_n,x_{n+1}}) \log (R^\theta_n(x_n,x_{n+1})\Delta t) 
\end{align*}

where on the last line we used $\log (1+z)=z-\frac{z^2}{2}+o\left(z^2\right)$ which is valid for $|z| \leq 1, z \neq-1$. For any finite $R_n^\theta\left(x_n, x_n\right), \Delta t$ can be taken small enough such that the series  expansion holds. Next we look at the expectation of this expression with respect to the distribution $q(x_{n+1}|x_{n},x_N)$:
\begin{align*}
&\mathbb{E}_{ x_{n+1}\sim q(x_{n+1}|x_{n},x_N)} [\log p_{\theta}(x_{n+1}|x_n )] =\\
&=\sum_{x_{n+1}}(\delta_{x_n,x_{n+1}} + R_n^q(x_n,x_{n+1}|x_N)\Delta t) \Big[\delta_{x_n,x_{n+1}}R_n^{\theta}(x_n,x_{n})\Delta t +\\
&\quad\quad\quad\quad\quad\quad\quad\quad\quad\quad\quad\quad\quad\quad\quad\quad\quad (1-\delta_{x_n,x_{n+1}}) \log (R_n^{\theta}(x_n,x_{n+1})\Delta t) \Big] \\
&= \delta_{x_n,x_{n+1}}(1+ R_n^q(x_n,x_{n+1}|x_N)\Delta t)R_n^{\theta}(x_n,x_{n})\Delta t +\\
&\quad\quad\quad\quad\quad\quad\quad\quad\quad\quad\quad\quad \sum_{x_{n+1}\neq x_n}R_n^q (x_n,x_{n+1}|x_N)\Delta t \;\log R_n^{\theta}(x_n,x_{n+1})\Delta t \\
&=R_n^{\theta}(x_n,x_{n})\Delta t + R_n^q(x_n,x_{n+1}|x_N)R_n^{\theta}(x_n,x_{n})(\Delta t)^2 +\\
&\quad\quad\quad\quad\quad\quad\quad\quad\quad\quad\quad\quad \sum_{x_{n+1}\neq x_n}R_n^q(x_n,x_{n+1}|x_N)\Delta t \;\log R_n^{\theta}(x_n,x_{n+1})\Delta t \\
&= R_n^{\theta}(x_n,x_{n})\Delta t + \sum_{x_{n+1}\neq x_n}R_n^q(x_n,x_{n+1}|x_N)\Delta t \;\log R_n^{\theta}(x_n,x_{n+1})\Delta t + o(\Delta t) \\
&= o(\Delta t) + \Delta t\sum_{x_{n+1}\neq x_n} R_n^q(x_n,x_{n+1}|x_N) \;\log R_n^{\theta}(x_n,x_{n+1}) \Delta t - R_n^{\theta}(x_n,x_{n+1})
\end{align*}


where $R_n^q(x_n,x_{n+1}|x_N)$ is the rate matrix associated with the transition kernel $q(x_{n+1}|x_{n},x_N)$. When considering a discrete approximation of continuous time, i.e. $\Delta t \rightarrow0$, $o(\Delta t)$ represents higher-order corrections (terms that vanish faster than $\Delta t$). Hence when considering the limit $\Delta t \rightarrow0$ these terms can be ignored, leading to 
\begin{align*}
    &\mathbb{E}_{ x_{n+1}\sim q(x_{n+1}|x_{n},x_N)} [\log p_{\theta}(x_{n+1}|x_n )] = \\
    &\Delta t\sum_{x_{n+1}\neq x_n}  R_n^q(x_n,x_{n+1}|x_N) \;\log R_n^{\theta}(x_n,x_{n+1}) \Delta t - R_n^{\theta}(x_n,x_{n+1})
\end{align*}

Now we use this expression to write:

\begin{align*}
    &\mathbb{E}_{ x_{n+1}\sim  q(x_{n+1}|x_{n},x_N)} \left[\log\frac{p_\theta(x_{n+1}|x_n )}{p_{\text{ref}}(x_{n+1}|x_n )} \right] \\
    &= \mathbb{E}_{ x_{n+1}\sim  q(x_{n+1}|x_{n},x_N)} [\log p_\theta(x_{n+1}|x_n )] - \mathbb{E}_{ x_{n+1}\sim  q(x_{n+1}|x_{n},x_N)} [p_{\text{ref}}(x_{n+1}|x_n )] \\   
    &= \Delta t \sum_{x_{n+1}\neq x_n} R_n^q(x_n,x_{n+1}|x_N)\;\log \frac{ R_n^{\theta}(x_n,x_{n+1})}{R_n^{\text{ref}}(x_n,x_{n+1})} + R_n^{\text{ref}}(x_n,x_{n+1}) - R_n^{\theta}(x_n,x_{n+1}) \\
\end{align*}

Plugging this expression into the DPO loss $ L_{\mathrm{DT}}(\theta)$ we get:

\begin{align*}
    &L_{\mathrm{DT}}(\theta) = -\log \sigma\Bigg[\beta \sum_{n=0}^N
    \mathbb{E}_{\substack{x_n^w\sim q(x_n|x_N^w) \\ x_n^l\sim q(x_n|x_N^l)}}\\
    &\Delta t\left(  \sum_{x_{n+1}\neq x_n^w}R_n^{\theta}(x_n^l,x_{n+1}|x_N) \;\log \frac{ R_n^{\theta}(x_n^w,x_{n+1})}{R_n^{\text{ref}}(x_n^w,x_{n+1})}+ R_n^{\text{ref}}(x_n^w,x_{n+1}) - R_n^{\theta}(x_n^w,x_{n+1})\right) \\
    - &\Delta t \left( \sum_{x_{n+1}\neq x_n^l}R_n^{\theta}(x_n^l,x_{n+1}|x_N) \;\log \frac{ R_n^{\theta}(x_n^l,x_{n+1})}{R_n^{\text{ref}}(x_n^l,x_{n+1})}+ R_n^{\text{ref}}(x_n^l,x_{n+1}) - R_n^{\theta}(x_n^l,x_{n+1})\right)\Bigg]
\end{align*}

Taking the limit of the discrete time loss $L_{\mathrm{DT}}(\theta)$ as  $N\rightarrow\infty$ (and hence $\Delta t = 1/N \rightarrow0$ ) we get back to the continuous time case:

\begin{align*}
    L_{\mathrm{CT}}(\theta) = &\lim_{\substack{N\to\infty \\\Delta t \rightarrow0}} L_{\mathrm{DT}}(\theta) = -\log \sigma\Bigg[\beta  
    \mathbb{E}_{\substack{x_n^w\sim q(x_n|x_N^w) \\ x_n^l\sim q(x_n|x_N^l)}} \int_0^1dt\\
    & \Bigg( \sum_{x_{n+1}\neq x_n^w}R_n^{\theta}(x_n^l,x_{n+1}|x_N) \;\log \frac{ R_n^{\theta}(x_n^w,x_{n+1})}{R_n^{\text{ref}}(x_n^w,x_{n+1})}+ R_n^{\text{ref}}(x_n^w,x_{n+1}) - R_n^{\theta}(x_n^w,x_{n+1}) \\
    &-  \sum_{x_{n+1}\neq x_n^l}R_n^{\theta}(x_n^l,x_{n+1}|x_N) \;\log \frac{ R_n^{\theta}(x_n^l,x_{n+1})}{R_n^{\text{ref}}(x_n^l,x_{n+1})}+ R_n^{\text{ref}}(x_n^l,x_{n+1}) - R_n^{\theta}(x_n^l,x_{n+1})\Bigg)\Bigg]
\end{align*}

We can estimate the integral with Monte Carlo if we consider it to be an expectation with respect to a uniform distribution over times $t \in [0, 1]$.
\begin{align*}
    L_{\mathrm{CT}}(\theta) = -\log \sigma\Bigg[&\beta \mathbb{E}_{\substack{t \sim \mathcal{U}[0,1]\\ x_t^w\sim q_{t|1}(x_t|x_1^w) \\ x_t^l\sim q_{t|1}(x_t|x_1^l)}}  \\
    &\Bigg( \sum_{j\neq x^w_t}R_t^{q}(x^w_t,j|x_1^w) \log \frac{ R_t^{\theta}(x^w_t,j)}{R_t^{\text{ref}}(x^w_t,j)}  + R_t^{\text{ref}}(x^w_t,j) - R_t^{\theta}(x^w_t,j)\\
    &- \sum_{j\neq x^l_t}R_t^{q}(x^l_t,j|x_1^l) \log \frac{ R_t^{\theta}(x^l_t,j)}{R_t^{\text{ref}}(x^l_t,j)} + R_t^{\text{ref}}(x^l_t,j)- R_t^{\theta}(x^l_t,j)\Bigg)\Bigg]
\end{align*}

Note that $-\log\sigma$ is a convex function and we can apply Jensen's inequality to yield:
\begin{align*}
    L_{\mathrm{CT}}(\theta) \leq -\mathbb{E}_{\substack{t \sim \mathcal{U}[0,1]\\ x_t^w\sim q_{t|1}(x_t|x_1^w) \\ x_t^l\sim q_{t|1}(x_t|x_1^l)}}
    \log \sigma\Bigg[ \beta\:\mathcal{D}^{\theta}_\text{ref}(x_t^w|x_1^w) - \beta\:\mathcal{D}_\text{ref}^{\theta}(x_t^l|x_1^l) \Bigg]
\end{align*}

where
\begin{align*}
    \mathcal{D}^{\theta}_\text{ref}(x_t|x_1) = \sum_{j\neq x_t}R_t^{q}(x_t,j|x_1) \log \frac{ R_t^{\theta}(x_t,j)}{R_t^{\text{ref}}(x_t,j)} + R_t^{\text{ref}}(x_t,j)- R_t^{\theta}(x_t,j)\quad .
\end{align*}

where $R_t^{q}(x_t,x_{t+dt}|x_1)$ depends on the chosen noise schedule and is defined as per \eqref{eq: cond rate matrix relu}, while $R_t^{\theta}(x_t,x_{t+dt})$ and $R_t^{\text{ref}}(x_t,x_{t+dt})$ are estimated as per \eqref{eq: rate matrix cond exp}.
\section{Multi-Dimensional \DthreePO{}} \label{sec: multidim dpo derivation}
In this section we adapt the \DthreePO{} loss to account for $D$-dimensional data. Consider $\boldsymbol{x}\in\{1,\cdots,S\}^D$ is a $D$-dimensional vector with components $x^d$ where $d=1,\dots,D$. We derive the DPO loss for this general case. The derivation proceeds in the same way as for the 1-dimensional case above, up to \eqref{eq: mid step derivation 1d}. For the $D$-dimensional case have:
\begin{align*}
    L_{\mathrm{DT}}(\theta) = -\log \sigma\Bigg(\beta N\mathbb{E}_n & \mathbb{E}_{\substack{\boldsymbol{x}_n^w\sim q(\boldsymbol{x}_n|\boldsymbol{x}_N^w)\\ \boldsymbol{x}_{n+1}^w\sim q(\boldsymbol{x}_{n+1}|\boldsymbol{x}_{n}^w, \boldsymbol{x}_N^w)}} \left[\log\frac{p_\theta(\boldsymbol{x}_{n+1}^w|\boldsymbol{x}_n^w )}{p_{\text{ref}}(\boldsymbol{x}_{n+1}^w|\boldsymbol{x}_n^w )} \right] \notag\\
    -& \mathbb{E}_{\substack{\boldsymbol{x}_n^l\sim q(\boldsymbol{x}_n|\boldsymbol{x}_N^l) \\  \boldsymbol{x}_{n+1}^l\sim q(\boldsymbol{x}_{n+1}|\boldsymbol{x}_{n}^l,\boldsymbol{x}_N^l)}} \left[\log \frac{p_\theta(\boldsymbol{x}_{n+1}^l|\boldsymbol{x}_n^l  )}{p_{\text{ref}}(\boldsymbol{x}_{n+1}^l|\boldsymbol{x}_n^l  )} \right]\Bigg) 
\end{align*}

In order to model transitions across multiple dimensions in a single time-step, we consider the following factorization of the transition probability:
\begin{align*}
    {p}_{\theta}\left(\boldsymbol{x}_{n+1} \mid \boldsymbol{x}_n\right)&=\prod_{d=1}^D {p}_{\theta}^d\left(x^d_{n+1} \mid \boldsymbol{x}_n\right) \,.
\end{align*}
By considering each dimension $x^d_{n+1}$ to be conditionally independent given the current vector $\boldsymbol{x}_n$, we can tractably account for multi-dimensional transitions in a single timestep. Similarly, we factorize 
\begin{align*}
    q\left(\boldsymbol{x}_{n+1} \mid \boldsymbol{x}_n,\boldsymbol{x}_N\right)&=\prod_{d=1}^D q^d\left(x^d_{n+1} \mid x_n^d,x_N^d\right) \, ,
\end{align*}
which aligns with the structure of the forward diffusion process, where noise is added independently across dimensions. Using this factorization we can rewrite the expectation terms as:
\begin{align*}
     \mathbb{E}_{\substack{\boldsymbol{x}_{n}\sim  q(\boldsymbol{x}_{n} \mid \boldsymbol{x}_N) \\\boldsymbol{x}_{n+1}\sim  q\left(\boldsymbol{x}_{n+1} \mid \boldsymbol{x}_n,\boldsymbol{x}_N\right)}} \left[\log\frac{p_\theta(\boldsymbol{x}_{n+1}|\boldsymbol{x}_n )}{p_{\text{ref}}(\boldsymbol{x}_{n+1}|\boldsymbol{x}_n )} \right] 
     = \sum_{d=1}^D\mathbb{E}_{\substack{\boldsymbol{x}_{n}\sim  q(\boldsymbol{x}_{n} \mid \boldsymbol{x}_N) \\x_{n+1}^d\sim  q^d\left(x^d_{n+1} \mid x_n^d,x_N^d\right)}} \left[\log\frac{p_\theta^d(x^d_{n+1}|\boldsymbol{x}_n )}{p_{\text{ref}}^d(x^d_{n+1}|\boldsymbol{x}_n )} \right]
\end{align*}

\textit{Proof}
\begin{align*}
     \mathbb{E}_{\substack{\boldsymbol{x}_{n}\sim  q(\boldsymbol{x}_{n} \mid \boldsymbol{x}_N) \\\boldsymbol{x}_{n+1}\sim  q\left(\boldsymbol{x}_{n+1} \mid \boldsymbol{x}_n,\boldsymbol{x}_N\right)}} \left[\log\frac{p_\theta(\boldsymbol{x}_{n+1}|\boldsymbol{x}_n )}{p_{\text{ref}}(\boldsymbol{x}_{n+1}|\boldsymbol{x}_n )} \right] 
     &= \mathbb{E}_{\substack{\boldsymbol{x}_{n}\sim  q(\boldsymbol{x}_{n} \mid \boldsymbol{x}_N) \\\boldsymbol{x}_{n+1}\sim  q\left(\boldsymbol{x}_{n+1} \mid \boldsymbol{x}_n,\boldsymbol{x}_N\right)}}  \left[\log\frac{\prod_{d=1}^D p_\theta^d(x^d_{n+1}|\boldsymbol{x}_n )}{\prod_{d=1}^Dp_{\text{ref}}^d(x^d_{n+1}|\boldsymbol{x}_n )} \right] \\
    &= \sum_{d=1}^D\mathbb{E}_{\substack{\boldsymbol{x}_{n}\sim  q(\boldsymbol{x}_{n} \mid \boldsymbol{x}_N) \\\boldsymbol{x}_{n+1}\sim  q\left(\boldsymbol{x}_{n+1} \mid \boldsymbol{x}_n,\boldsymbol{x}_N\right)}}  \left[\log\frac{p_\theta^d(x^d_{n+1}|\boldsymbol{x}_n )}{p_{\text{ref}}^d(x^d_{n+1}|\boldsymbol{x}_n )} \right] \\
    &     = \sum_{d=1}^D\mathbb{E}_{\substack{\boldsymbol{x}_{n}\sim  q(\boldsymbol{x}_{n} \mid \boldsymbol{x}_N) \\x_{n+1}^d\sim  q^d\left(x^d_{n+1} \mid x_n^d,x_N^d\right)}} \left[\log\frac{p_\theta^d(x^d_{n+1}|\boldsymbol{x}_n )}{p_{\text{ref}}^d(x^d_{n+1}|\boldsymbol{x}_n )} \right]
\end{align*}

\textit{Where on the last line we use the fact that the term inside the expectation depends on $\boldsymbol{x}_{n+1}$ only via its $d$-dimensional component ${x}^d_{n+1}$.} 

Substituting this expression into the DPO loss we get:
\begin{align*}
    L_{\mathrm{DT}}(\theta) = -\log \sigma\Bigg(\beta N\mathbb{E}_n \sum_{d=1}^D \:& \mathbb{E}_{\boldsymbol{x}_n^w\sim q(\boldsymbol{x}_n|\boldsymbol{x}_N^w)}\mathbb{E}_{x_{n+1}^d\sim  q^d\left(x^d_{n+1} \mid x_n^d,x_N^{d,w}\right)} \left[\log\frac{p_\theta^d(x^d_{n+1}|\boldsymbol{x}_n^w )}{p_{\text{ref}}^d(x^d_{n+1}|\boldsymbol{x}_n^w )} \right] \\
    -& \mathbb{E}_{\boldsymbol{x}_n^l\sim q(\boldsymbol{x}_n|\boldsymbol{x}_N^l)} \mathbb{E}_{x_{n+1}^d\sim  q^d\left(x^d_{n+1} \mid x_n^d,x_N^{d,l}\right)} \left[\log\frac{p_\theta^d(x^d_{n+1}|\boldsymbol{x}_n^l )}{p_{\text{ref}}^d(x^d_{n+1}|\boldsymbol{x}_n^l )} \right] \Bigg)
\end{align*}

We can now follow the same derivation steps as for the 1-dimensional case, leading to:
\begin{align*}
    L_{\mathrm{CT}}(\theta) = -\mathbb{E}_{t\sim\mathcal{U}(0,1), \boldsymbol{x}^w_t\sim q(\boldsymbol{x}_t|\boldsymbol{x}_1^w), \boldsymbol{x}^l_t\sim q(\boldsymbol{x}_t|\boldsymbol{x}_1^l)}\log \sigma\Bigg[\beta\sum_{d=1}^D  \left(\mathcal{D}_{\text{ref}}^{\theta,d}(\boldsymbol{x}_t^w|\boldsymbol{x}_1^w) - \mathcal{D}_{\text{ref}}^{\theta,d}(\boldsymbol{x}_t^l|\boldsymbol{x}_1^l) \right) \Bigg]
\end{align*}

where
\begin{align}
    \mathcal{D}_{\text{ref}}^{\theta,d}(\boldsymbol{x}_t|\boldsymbol{x}_1) = \sum_{j^d\neq x^d_t}R_t^{d,q}(x_t^d,j^d | x_1^d) \log \frac{ R_t^{d,\theta}(\boldsymbol{x}_t,j^d)}{R_t^{d,\text{ref}}(\boldsymbol{x}_t,j^d)}  + R_t^{d,\text{ref}}(\boldsymbol{x}_t,j^d) -R_t^{d,\theta}(\boldsymbol{x}_t,j^d)
    \label{eq: bigD}
\end{align}

where $R_t^{d,\theta}(\boldsymbol{x},j^d) = \mathbb{E}_{p^{d,\theta}_{1|t}(x_1^d|\boldsymbol{x})}[R_t^{d,\theta}(x^d,j^d|x_1^d)]$, and $x^d$ denotes the $d$-dimensional component of vector $\boldsymbol{x}$.
\section{\DthreePO{} Loss for Masking State Models} \label{sec: dpo for masking state}
In this section we adapt the \DthreePO{} loss for the specific case of masking noise process. In $D$ dimensions we consider independent corruption processes in each dimension, similar to the factorization assumptions made in continuous diffusion models where the forward noising processes proceed independently in each dimension.
\begin{align*}
    q^{\text{mask}}_{t \mid 1}\left(\boldsymbol{x}_t \mid \boldsymbol{x}_1\right) & =\prod_{d=1}^D q^{\text{mask},d}_{t \mid 1}\left(x_t^d \mid x_1^d\right) \\
    & =\prod_{d=1}^D\left(t \delta\left\{x_t^d, x_1^d\right\}+(1-t) \delta\left\{x_t^d, M\right\}\right)
\end{align*}

In this case, the conditional rate matrix for the masking process can be derived in closed form as:
\begin{align}
    R_t^{q,d}\left(x_t^d, x_{t+dt}^d \mid x_1^d\right)&=\frac{\operatorname{ReLU}\left(\partial_t q^{\text{mask},d}_{t \mid 1}\left(x_{t+dt}^d \mid x_1^d\right)-\partial_t q^{\text{mask},d}_{t \mid 1}\left(x_t^d \mid x_1^d\right)\right)}{S \cdot q^{\text{mask},d}_{t \mid 1}\left(x_t^d \mid x_1^d\right)} \notag\\
    &=\frac{1}{1-t} \delta\left\{x_t^d, M\right\} \delta\left\{x^d_{t+dt}, x^d_1\right\}
    \label{eq: cond rate matrix mask}
\end{align}

We can then express the unconditional rate matrix as:
\begin{align}
    R_t^{d,\theta}(\boldsymbol{x}_t, x_{t+dt}^d) &= \mathbb{E}_{p_{1 \mid t}^{\theta}(x_1^d \mid \boldsymbol{x}t)}\left[R_t^{\text{mask},d}\left(x_t^d, x_{t+dt}^d \mid x_1^d\right)\right] \notag\\
    &= \mathbb{E}_{p_{1 \mid t}^{\theta}(x_1^d \mid \boldsymbol{x}t)}\left[\frac{1}{1-t} \delta\left\{x_t^d, M\right\} \delta\left\{x^d_{t+dt}, x^d_1\right\}\right] \notag\\
    &= \frac{1}{1-t} \delta\left\{x_t^d, M\right\} p_{1 \mid t}^\theta\left(x_1^d=x_{t+dt}^d \mid \boldsymbol{x}_t\right)
    \label{eq: r theta mask}
\end{align}

which vanishes for $x_t^d\neq M$ and for $x_{t+dt}^d=M$ as $p_{1 \mid t}^\theta\left(x_1^d=M \mid \boldsymbol{x}_t\right) = 0$, meaning $\boldsymbol{x}_1$ cannot have any masked dimensions. Substituting \eqref{eq: cond rate matrix mask} and \eqref{eq: r theta mask} into \eqref{eq: bigD}:

\begin{align}
    \mathcal{D}^{\theta,d}(\boldsymbol{x}_t|\boldsymbol{x}_1) &= \sum_{j^d\neq x^d_t}R_t^{d,q}(x_t^d,j^d | x_1^d) \log \frac{ R_t^{d,\theta}(\boldsymbol{x}_t,j^d)}{R_t^{d,\text{ref}}(\boldsymbol{x}_t,j^d)}  + R_t^{d,\text{ref}}(\boldsymbol{x}_t,j^d) -R_t^{d,\theta}(\boldsymbol{x}_t,j^d) \notag\\
    &= \frac{\delta\{x^d_t, M\}}{1-t} \sum_{j^d\neq M} \delta\{x^d_1, j^d\} \log\frac{p_{1 \mid t}^\theta\left(j^d \mid \boldsymbol{x}_t\right)}{p_{1 \mid t}^{\text{ref}}\left(j^d \mid \boldsymbol{x}_t\right)} +p_{1 \mid t}^{\text{ref}}\left(j^d \mid \boldsymbol{x}_t\right) -p_{1 \mid t}^\theta\left(j^d \mid \boldsymbol{x}_t\right) \notag\\
    &=\frac{\delta\{x^d_t, M\}}{1-t} \log\frac{p_{1 \mid t}^\theta\left(x_1^d \mid \boldsymbol{x}_t\right)}{p_{1 \mid t}^{\text{ref}}\left(x_1^d \mid \boldsymbol{x}_t\right)}
    \label{eq: D noiseless}
\end{align}

Where on the last line we use the fact that the neural network $p_{1 \mid t}^\theta\left(\cdot \mid \boldsymbol{x}_t\right)$ outputs a probability distribution over all unmasked tokens to write $\sum_{j^d\neq M}p_{1 \mid t}^{\text{ref}}\left(x_1^d=j^d \mid \boldsymbol{x}_t\right)=\sum_{j^d\neq M}p_{1 \mid t}^\theta\left(x_1^d=j^d \mid \boldsymbol{x}_t\right)=1$. Hence the final loss is:
\begin{align*}
    L^{\text{mask}}_{\mathrm{CT}}(\theta) &= -\mathbb{E}_{\substack{t\sim\mathcal{U}[0,1] \\ \boldsymbol{x}_t^w\sim q(\boldsymbol{x}_t|\boldsymbol{x}_1^w)\\ \boldsymbol{x}_t^l\sim q(\boldsymbol{x}_t|\boldsymbol{x}_1^l)}}\log \sigma\Bigg[\frac{\beta}{1-t}\sum_{d=1}^D \\
    &\left(  \delta\{x^{d,w}_t, M\} \log\frac{p_{1 \mid t}^\theta\left(x_1^{d,w} \mid \boldsymbol{x}_t^w\right)}{p_{1 \mid t}^{\text{ref}}\left(x_1^{d,w} \mid \boldsymbol{x}_t^w\right)} 
    -\delta\{x^{d,l}_t, M\}\log\frac{p_{1 \mid t}^\theta\left(x_1^{d,l} \mid \boldsymbol{x}_t^l\right)}{p_{1 \mid t}^{\text{ref}}\left(x_1^{d,l} \mid \boldsymbol{x}_t^l\right)}  \right) \Bigg]
\end{align*}

Similar to the classical DPO loss in \eqref{eq:DPOloss}, this loss is based on the difference in log probabilities assigned to recovering the original samples under the learned model $p^\theta_{1|t}$ compared to a reference model $p^{\text{ref}}_{1|t}$. However, this difference is weighted by a masking indicator, ensuring that only masked dimensions contribute to the loss. Intuitively, the effect of optimizing this objective is to increase the model’s likelihood of reconstructing the preferred sample $x^w$ while reducing the likelihood of reconstructing the dis-preferred sample $x^l$, making $x^w$ more likely to be recovered during the unmasking process.

\subsection{Masking with Additional Uniform Noise} \label{sec: uniform noise masking}
We now consider the case in which we introduce a non-zero probability to transition from an unmasked state back to a masked state during the denoising process. Intuitively this allows more flexibility at inference time as the model could potentially recover from errors by re-masking certain tokes. \cite{campbell2024flow} show that such an additional noise process is in detailed balance with the noise-free process and hence does not affect the final data distribution at time $t=1$. They also show that the resulting rate matrix for a noise process with coefficient $\eta$ is given by:

\begin{align*}
    R_t^{d,\theta}(\boldsymbol{x}_t,j^d)&=\frac{1+\eta t}{1-t} p_{1 \mid t}^\theta\left(x_1^d=j^d \mid \boldsymbol{x}_t\right) \delta\left\{x_t^d, M\right\}+\eta\left(1-\delta\left\{x_t^d, M\right\}\right) \delta\left\{j^d, M\right\}
    \\&=
    \begin{cases}
        \frac{1+\eta t}{1-t}p_{1 \mid t}^\theta\left(x_1^d=j^d \mid \boldsymbol{x}_t\right) &\text{ for } x_t^d= M,j^d\neq M\\
        \eta & \text{ for } x_t^d\neq M,j^d= M \\
        0 & \text{otherwise}
    \end{cases}
\end{align*}

While $R_t^{d,q}(x_t^d,x_{t+dt}^d | x_1^d)$ remains unaffected. Substituting this into \eqref{eq: bigD}:
\begin{align}
    \mathcal{D}^{\theta,d}_t(\boldsymbol{x}) &= \sum_{j^d\neq x^d}R_t^{d,q}(x_t^d,j^d | x_1^d) \log \frac{ R_t^{d,\theta}(\boldsymbol{x},j^d)}{R_t^{d,\text{ref}}(\boldsymbol{x},j^d)}  + R_t^{d,\text{ref}}(\boldsymbol{x},j^d) -R_t^{d,\theta}(\boldsymbol{x},j^d) \notag\\
    &=\frac{1+\eta t}{1-t}\delta\{x^d_t, M\} \log\frac{p_{1 \mid t}^\theta\left(x_1^d \mid \boldsymbol{x}_t\right)}{p_{1 \mid t}^{\text{ref}}\left(x_1^d \mid \boldsymbol{x}_t\right)} + \left(1-\delta\left\{x_t^d, M\right\}\right) \delta\left\{j^d, M\right\} \left(\eta\log \frac{\eta}{\eta}+\eta-\eta\right) \notag\\
    &=\frac{1+\eta t}{1-t}\delta\{x^d_t, M\} \log\frac{p_{1 \mid t}^\theta\left(x_1^d \mid \boldsymbol{x}_t\right)}{p_{1 \mid t}^{\text{ref}}\left(x_1^d \mid \boldsymbol{x}_t\right)}
    \label{eq: D noisy}
\end{align}

which is the same as for the noiseless reverse process, up to a multiplicative constant $1+\eta t$. Hence the final loss is:
\begin{align*}
    L^{\text{mask}}_{\mathrm{CT}}(\theta) &= -\mathbb{E}_{\substack{t\sim\mathcal{U}[0,1] \\ \boldsymbol{x}_t^w\sim q(\boldsymbol{x}_t|\boldsymbol{x}_1^w)\\ \boldsymbol{x}_t^l\sim q(\boldsymbol{x}_t|\boldsymbol{x}_1^l)}}\log \sigma\Bigg[\frac{\beta(1+\eta t)}{1-t}\sum_{d=1}^D \\
    &\left(  \delta\{x^{d,w}_t, M\} \log\frac{p_{1 \mid t}^\theta\left(x_1^{d,w} \mid \boldsymbol{x}_t^w\right)}{p_{1 \mid t}^{\text{ref}}\left(x_1^{d,w} \mid \boldsymbol{x}_t^w\right)} 
    -\delta\{x^{d,l}_t, M\}\log\frac{p_{1 \mid t}^\theta\left(x_1^{d,l} \mid \boldsymbol{x}_t^l\right)}{p_{1 \mid t}^{\text{ref}}\left(x_1^{d,l} \mid \boldsymbol{x}_t^l\right)}  \right) \Bigg]
\end{align*}

\subsection{Complexity Analysis for Masking Noise Process} \label{sec: complexity masking}
For the masking noise process, the derived expressions for $\mathcal{D}^{\theta,d}_t(\boldsymbol{x})$ in Equations \eqref{eq: D noiseless} and \eqref{eq: D noisy} provide a computationally efficient way to estimate the \DthreePO{} loss function. In practice, the denoising models $p_{1 \mid t}^{\theta}$ and $p_{1 \mid t}^{\text{ref}}$ take as input a noisy vector $\boldsymbol{x}_t \in \{1,\dots,S,M\}^D$ and output probability vectors $p_{1 \mid t}(\boldsymbol{x}_1 \mid \boldsymbol{x}_t) \in [0,1]^D$. Since the loss function requires evaluating the probability of reconstructing each dimension $x_1^d$, this can be directly accessed as the $d^{\text{th}}$ component of the model’s output.

Due to the structure of the masking noise process, computing the sum $\sum_{d=1}^D \mathcal{D}^{\theta,d}_t(\boldsymbol{x})$ is particularly efficient. The required probability vectors $p_{1 \mid t}^{\theta}(\boldsymbol{x}_1 \mid \boldsymbol{x}_t)$ and $p_{1 \mid t}^{\text{ref}}(\boldsymbol{x}_1 \mid \boldsymbol{x}_t)$ can be obtained with a single forward pass for each model. As a result, evaluating $\sum_{d=1}^D \mathcal{D}^{\theta,d}_t(\boldsymbol{x})$ requires exactly two model queries: one for the learned model $p_{1 \mid t}^{\theta}$ and one for the reference model $p_{1 \mid t}^{\text{ref}}$.

When estimating the \textit{per-example} \DthreePO{} loss using a batch of size $T$ to approximate the expectation over $t \sim \mathcal{U}[0,1]$, the total number of model queries scales to $2T = O(T)$. For a dataset containing $P$ preference pairs, the overall computational complexity becomes $O(PT)$, reflecting a linear dependence on both the number of preferences and batch size. This scaling ensures that preference optimization in discrete diffusion models remains computationally efficient, making it practical for large-scale generative modeling tasks.

\end{document}